\documentclass[10pt,twocolumn,letterpaper]{article}

\usepackage{iccv}
\usepackage{times}
\usepackage{epsfig}
\usepackage{graphicx}
\usepackage{amsmath}
\usepackage{amssymb}

\usepackage{booktabs}
\usepackage[linesnumbered,ruled]{algorithm2e}
\usepackage{amsmath}
\usepackage{subcaption}
\usepackage{pifont}% http://ctan.org/pkg/pifont
\usepackage{enumitem}
\usepackage{multicol}
\usepackage{multirow}
\usepackage{mathtools}
\usepackage{xcolor}

\usepackage[pagebackref=true,breaklinks=true,letterpaper=true,colorlinks,bookmarks=false]{hyperref}

\iccvfinalcopy % *** Uncomment this line for the final submission

\ificcvfinal\fi

\begin{document}

%%%%%%%%% TITLE
\title{Monocular 3D Object Detection with Bounding Box Denoising in 3D by Perceiver}

\author{Xianpeng Liu$^{1}$\thanks{Work partially conducted during an internship at OPPO Seattle Research Center, USA.} , Ce Zheng$^{2}$, Kelvin Cheng$^{1}$, Nan Xue$^{3}$, Guo-Jun Qi$^{4,5}$, Tianfu Wu$^{1}$\\
$^1$North Carolina State University \quad $^2$University of Central Florida \quad $^3$Wuhan University \\
$^4$OPPO Seattle Research Center, USA \quad $^5$Westlake University\\
{\tt\small \{xliu59, kbcheng, tianfu\_wu\}@ncsu.edu; cezheng@knights.ucf.edu;} \\{\tt\small xuenan@ieee.org; guojunq@gmail.com}\\
}

\maketitle

\ificcvfinal\thispagestyle{empty}\fi

%%%%%%%%% ABSTRACT
\begin{abstract}
The main challenge of monocular 3D object detection is the accurate localization of 3D center. Motivated by a new and strong observation that this challenge can be remedied by a 3D-space local-grid search scheme in an ideal case, we propose a stage-wise approach, which combines the information flow from 2D-to-3D (3D bounding box proposal generation with a single 2D image) and 3D-to-2D (proposal verification by denoising with 3D-to-2D contexts) in a top-down manner. Specifically, we first obtain initial proposals from off-the-shelf backbone monocular 3D detectors. Then, we generate a 3D anchor space by local-grid sampling from the initial proposals. Finally, we perform 3D bounding box denoising at the 3D-to-2D proposal verification stage. To effectively learn discriminative features for denoising highly overlapped proposals, this paper presents a method of using the Perceiver I/O model~\cite{perceiverio} to fuse the 3D-to-2D geometric information and the 2D appearance information. With the encoded latent representation of a proposal, the verification head is implemented with a self-attention module. Our method, named as \textbf{MonoXiver}, is generic and can be easily adapted to any backbone monocular 3D detectors. Experimental results on the well-established KITTI dataset and the challenging large-scale Waymo dataset show that MonoXiver consistently achieves improvement with limited computation overhead.
\end{abstract}

%%%%%%%%% BODY TEXT

\section{Introduction}

Detecting and locating objects in 3D using a single image, known as monocular 3D object detection, is a highly challenging task due to its ill-posed nature. However, the practical applications of low-cost system setups in fields such as self-driving cars and robotic manipulation have led to the development of robust monocular 3D object detection systems, making it a prominent research topic in the computer vision community.

Although significant progress has been achieved in this field~\cite{m3dsurvey}, accurate 3D center localization remains a major challenge for state-of-the-art (SOTA) methods as indicated in~\cite{monodle}. Most of these methods follow a bottom-up paradigm, where a single 2D image is used to directly predict 3D bounding boxes (2D-to-3D) with or without extra information (e.g. LiDAR). However, due to the inherent depth ambiguity of this task, relying solely on bottom-up 2D-to-3D paradigms may not be enough to completely address this challenge. 

In this paper, we observe that bottom-up 2D-to-3D predicted 3D bounding boxes are able to provide informative priors for monocular 3D object detection. We propose to leverage these contexts to improve detection performance in a top-down manner. Our proposed approach is motivated by a strong empirical upper-bound analysis with SOTA bottom-up monocular 3D object detectors.

\begin{table}[t]
\begin{center}
% \vspace{2mm}
\resizebox{0.99\linewidth}{!}{
    \begin{tabular}{c|c|ccc|c|c}
    \toprule
    Range & Stride & Easy & Moderate & Hard & \#Bboxes & mean IoU$_{avg}^5$\\
    \midrule
    
    \multicolumn{2}{c|}{MonoCon~\cite{monocon}}  & 26.33 & 19.01 & 15.98 & n & -\\
    
    \midrule
    
    $\pm$ 2 & 0.1 & 76.98 \textcolor{blue}{$_\uparrow$50.65} & 64.93 \textcolor{blue}{$_\uparrow$45.92} & 56.91 \textcolor{blue}{$_\uparrow$40.93} & 1600$\cdot$n & 0.931 \\
    $\pm$ 1.5 & 0.1 & 76.45  \textcolor{blue}{$_\uparrow$50.12} & 62.27  \textcolor{blue}{$_\uparrow$43.26} & 54.34  \textcolor{blue}{$_\uparrow$38.36} & 900$\cdot$n & 0.931 \\
    $\pm$ 1.5 & 0.2 & 74.00  \textcolor{blue}{$_\uparrow$47.67} & 61.78  \textcolor{blue}{$_\uparrow$42.77} & 53.64  \textcolor{blue}{$_\uparrow$37.66} & 225$\cdot$n & 0.868\\
    $\pm$ 1.5 & 0.3 & 64.80  \textcolor{blue}{$_\uparrow$38.47} & 56.20  \textcolor{blue}{$_\uparrow$37.19} & 50.66  \textcolor{blue}{$_\uparrow$34.68} & 121$\cdot$n & 0.812 \\
    $\pm$ 1.5 & 0.5 & 54.00  \textcolor{blue}{$_\uparrow$27.67} & 45.93  \textcolor{blue}{$_\uparrow$26.92} & 40.99  \textcolor{blue}{$_\uparrow$25.01} & 49$\cdot$n & 0.714 \\
    $\pm$ 1.5 & 0.75 & 41.58  \textcolor{blue}{$_\uparrow$15.25} & 34.44  \textcolor{blue}{$_\uparrow$15.43} & 30.12  \textcolor{blue}{$_\uparrow$14.14} & 25$\cdot$n & 0.612 \\
    
    \bottomrule
    \end{tabular}
}
\end{center} \vspace{-4mm}
\caption{
\textbf{A strong empirical upper-bound analysis.} Our MonoXiver is motivated by the observation that bottom-up monocular 3D object detectors can be significantly improved by leveraging a simple 3D-space local-grid search scheme in an ideal case. This highlights the potential of exploring the 3D proposal space. However, this improvement comes at a cost: 1) the proposal verification stage experiences significant increases in number of bounding box proposals, and 2) a more powerful refinement module is required to handle highly overlapped boxes.
}\label{Tab:upper_bound}
  \vspace{-2mm}
\end{table}

\noindent \textbf{The empirical upper-bound experiment.} We analyze vehicle detection performance on the well-established KITTI validation set \cite{mono3d, chen2, multiview}. We start with the recently proposed MonoCon~\cite{monocon} because it achieves SOTA performance with a very simple design. By using MonoCon's prediction results as bottom-up 3D anchors, we sample 3D bounding box proposals along both the $x$ and $z$ axes (i.e., in the bird-eye's view) with a simple strategy. First, we define a 2D local grid with a range (e.g., $2$ meter) and a stride (e.g., $0.1$ meter). For each bottom-up anchor, with its center located at the local-grid origin, we replicate the anchor at each grid vertex without changing its size, orientation and prediction score. We use all the generated 3D bounding boxes as top-down proposals and compute the empirical upper-bound of 3D detection performance by searching for the best match within top-down proposals for each ground-truth 3D object bounding box based on 3D Intersection-over-Union (3D IoU).

\textbf{A strong observation from the experiment.} As shown in Table~\ref{Tab:upper_bound}, despite the evaluation result of initial bottom-up proposals is relatively low in 3D average precision ($AP_{3D}$) (e.g., 19.01 in Moderate settings), \textit{most of predicted 3D centers are already in the close proximity of the GT ones.} \textbf{An empirical upper-bound of 34.4 $AP_{3D}$ can be achieved even with a coarse sampling scheme} (e.g., the last row), which is significantly higher than SOTA methods. We also obtained similar observations from other backbone 3D object detectors such as the MonoDLE~\cite{monodle} and the SMOKE~\cite{smoke}. These strong empirical observations demonstrate the potential of integrating the bottom-up initial proposals with top-down sampling and verification.

\begin{figure*}
    \centering
    \includegraphics[width=0.9\textwidth]{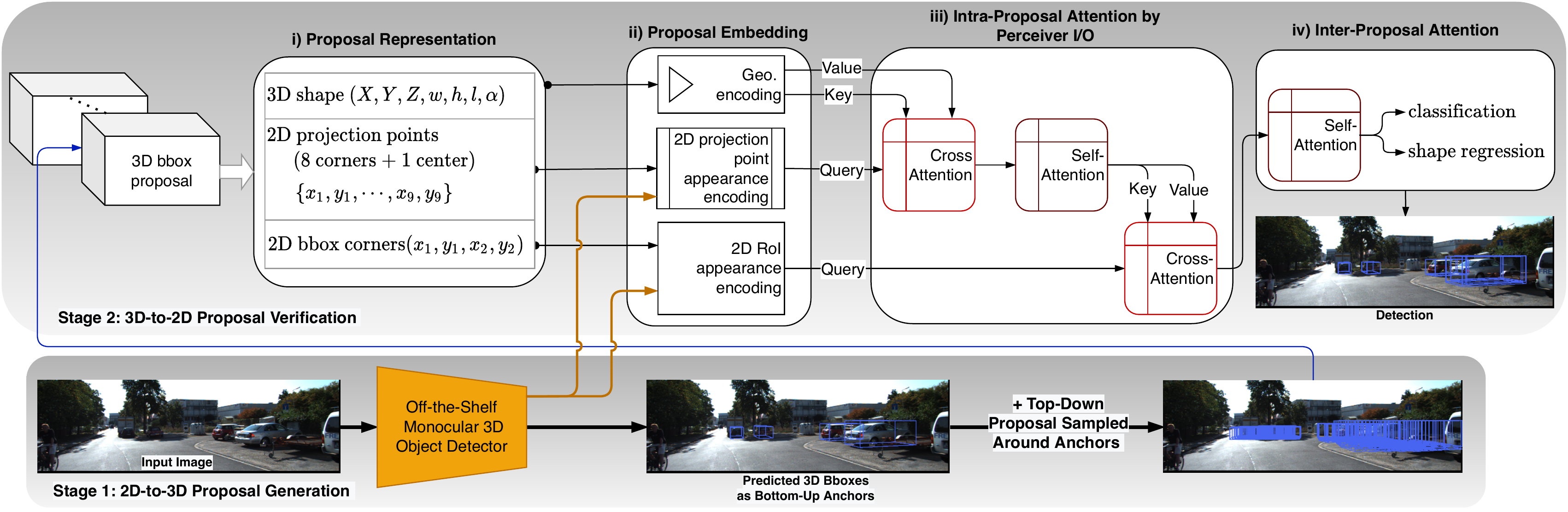}
    \caption{\textbf{Illustration of the proposed MonoXiver method.} MonoXiver is built for any off-the-shelf backbone monocular 3D object detectors. It consists of a 2D-to-3D proposal generation phase using a bottom-up anchoring and top-down sampling strategy, and a 3D-to-2D proposal verification (or denoising) phase using the Perceiver I/O~\cite{perceiverio} with a carefully designed 3D/2D input space to address the unique challenges. See text for details.}
    \label{fig:monoxiver}
    \vspace{-2mm}
\end{figure*}

\textbf{The challenge in the 3D-to-2D proposal verification.} The 3D-to-2D proposal verification phase can be treated as a 3D bounding box denoising process, as we want to search for the ``best" bounding boxes from the top-down proposal set. This process is extremely challenging, as the proposals generated from the same bottom-up anchor are highly overlapped in both 3D and 2D (after projection), which leads to the long-standing problem of handling the ``crowd" in detection. To quantitatively analyze the overlap extent after projection for top-down proposals, we consider one proposal and its top-$k$ overlapping neighbors in terms of Intersection-over-Union (IoU), which is denoted by IoU$_{avg}^k$ as the average IoU over the top-$k$ neighbors. The last column in Table~\ref{Tab:upper_bound} shows the statistics. 

The statistics clearly demonstrates the difficulty of denoising densely generated top-down proposal set (e.g., stride=0.1, IoU$_{avg}^5$=0.931). Even for a relative sparse generated top-down proposal set (e.g., stride=0.75, IoU$_{avg}^5$=0.612), the challenge still exists because proposals sampled in front and behind of the same anchor will almost collapse to the same 2D bounding box, especially when they are far away from the camera. So, how can we encode the 3D bounding box proposal for verification under the monocular setting? From this perspective, we note that many methods that work well in multi-view 3D object detection are often not applicable for monocular 3D objection because they mainly rely on features that are obtained by fusing projection features from multi-view feature maps. 

Based on the intuition that even though highly overlapped proposals have similar appearance features, their inherent 3D-to-2D geometric features (e.g. 3D location, projected geometry, etc.) are extremely different, we propose to fuse these 3D-to-2D geometric feature with their correspondent appearance features to learn discriminative features for bounding box denoising. We present a method of using the Perceiver I/O model~\cite{perceiverio} to effectively fuse these contexts, as Perceiver has shown strong capabilities to fuse multi-modal inputs. With the encoded latent representation of a proposal, the verification head is implemented by a self-attention module (see the top in Fig.~\ref{fig:monoxiver}). The proposed method is named as \textbf{MonoXiver}, indicating its general applicability to any backbone monocular 3D detectors and the integration of the Perceiver model.

In experiments, we evaluate our proposed MonoXiver on the well-established KITTI benchmark \cite{kitti} and the challenging large-scale Waymo \cite{waymo} dataset with various backbone monocular 3D detectors. It achieves consistent and significant performance improvement on both datasets with limited computation overhead. Moreover, it achieves the 1st place among monocular methods on the KITTI vehicle detection benchmark, outperforming the previous works by a large margin.

\section{Related Works and Our Contributions}

\noindent\textbf{Monocular 3D Object Detection.} The problem of monocular 3D object detection is considered ill-posed, prompting recent research efforts to leverage additional information sources, including LiDAR point clouds~\cite{monorun, caddn, monodistill, monojsg, monodtr}, depth estimation~\cite{patchnet, d4lcn, demystifying, ddmp3d, pct, dfrnet, dd3d}, CAD models~\cite{autoshape, dcd}, temporal frames~\cite{kinematic3d}, etc. These approaches have shown improved detection performance when compared to purely image-based methods~\cite{3dvp, deep3dbox, smoke, m3drpn}. However, they often come with increased computation load and inference time, making them less practical for real-time applications such as autonomous vehicles. In contrast, purely image-based methods have seen significant performance improvements in the literature by exploiting geometric constraints~\cite{deep3dbox, monoflex, gupnet, monorcnn, monogeo, monodde}, perspective projections~\cite{rtm3d, am3d}, auxiliary training \cite{monocon}, novel loss designs~\cite{monopair, monodle, monodis} and second-stage processing techniques~\cite{nms3d}. These works focus on developing better bottom-up paradigms. Our proposed method, MonoXiver, takes a top-down paradigm, which explores 3D space differently than the aforementioned methods. It provides a generic and highly efficient second-stage refinement module for any pretrained state-of-the-art monocular 3D object detection method, making it a complementary approach. \\
\textbf{Bird's-Eye-View 3D Object Detection. }Bird's-Eye-View (BEV) 3D object detection \cite{bevsurvey} has recently gained significant attention as it provides a physical-interpretable feature space for integrating data from diverse sensors \cite{detr3d, petr, bevformer, lss} and modalities \cite{spatialdetr, bev4d}. To enable the learning of a physical-interpretable feature space, accurate physical measurements from depth perception systems such as multi-view cameras and LiDAR are crucial \cite{liga}. However, detecting 3D objects in BEV is challenging for monocular systems due to the intrinsic depth ambiguity. Directly accumulating image features based on projection reference points similar to recent query-based multi-view 3D detectors \cite{detr3d, petr} might fail because of lacking stereo-matching mechanism to resolve the depth ambiguity. To overcome this issue, recent BEV-based monocular methods \cite{caddn, pseudostereo, dfm} lift the image feature space to voxel-represented world-space using monocular depth estimation models. Our proposed method, MonoXiver, samples top-down proposals based on bottom-up initial proposals in BEV space. Unlike prior arts \cite{caddn, pseudostereo, dfm} that extract features represented in world-space, MonoXiver's top-down proposals extract features from both projected appearance features and inherent geometric features. We explore to address the challenging intrinsic feature ambiguity problem by applying a Perceiver-like attention mechanism to query informative features for 3D bounding box verification and refinement without the need for dense depth supervision. \\
\textbf{Transformers for Multi-modal Data.} Transformers \cite{attention-is-all-you-need}, known for their powerful attention mechanism, have been successfully applied in multiple research areas and applications \cite{bert, detr, vlbert, swin}. With the powerful attention mechanism, transformers can conveniently align multi-modal structured data, enabling the generation of universal multi-modal representation. The recent proposed Perceiver \cite{perceiver, perceiverio} extends the transformers' querying mechanism and makes it capable of efficiently handling data from arbitrary settings. In our work, we adapt Perceiver to fuse appearance features and geometric features from the same bounding box to generate appearance-geometric aligned representation, which is then used as queries for the down-stream 3D bounding box refinement task.

\noindent \textbf{Our Contributions.} MonoXiver makes three main contributions to the field of monocular 3D object detection:
\begin{itemize} [leftmargin=*]
\itemsep 0em 
    \item We conduct a new and strong empirical upper-bound analysis for state-of-the-art monocular 3D object detectors. We demonstrate that a simple, universally applicable sampling strategy can lead to consistent and significant performance improvement in an ideal situation. Based on these observations, we propose to explore the top-down, stage-wise detection paradigm with 2D-to-3D proposal generation and 3D-to-2D proposal verification for monocular 3D object detection. 
    
    \item To address the challenge at the bounding box denoising stage, we propose to learn discriminative features by enhancing appearance features with 3D-to-2D inherent geometric features. We therefore propose a carefully-designed module with the integration of the powerful Perceiver and self-attention mechanism. The resulting module can be applied generically to any off-the-shelf monocular 3D object detectors. 

    \item Extensive experiments demonstrate the effectiveness and efficiency of our proposed method, MonoXiver, under a separate two-stage training setting. It achieves consistent improvement with limited overhead on both KITTI and Waymo dataset. It achieves the 1st place among monocular methods on the KITTI vehicle detection benchmark, outperforming the previous works by a large margin.
\end{itemize}

\section{Approach}

In this section, we first present the straightforward 2D-to-3D proposal generation on top of a backbone monocular 3D object detector. Then, we present the details of the proposed MonoXiver method (Fig.~\ref{fig:monoxiver}).

\subsection{The 2D-to-3D Proposal Generation}
% Let $I$ be an RGB input image defined on the domain $\Lambda$, the goal of monocular 3D object detection is to detect the 3D bounding boxes with their class labels, denoted by $\ell$ for each object instance in $I$.
% The 3D bounding box is parameterized by the 3D center position $\mathbf{P} = (X,Y,Z)$ in meters, the shape dimensions $\mathbf{D} = ({h}, {w}, {l})$ in meters and the observation angle $\alpha \in [-\pi, \pi] $, all measured in the camera coordinate system. 
Let $I$ be an RGB input image defined on the domain $\Lambda$. The goal of monocular 3D object detection is to detect 3D bounding boxes, along with their class labels denoted by $\ell$ for each object instance in $I$. The 3D bounding box is parameterized by the 3D center position $\mathbf{P} = (X,Y,Z)$ in meters, the shape dimensions $\mathbf{D} = ({h}, {w}, {l})$ in meters, and the observation angle $\alpha \in [-\pi, \pi]$, all measured in the camera coordinate system.

For simplicity, consider the pure image-based settings, a monocular 3D object detector often consists of two main components in inference: the feature backbone (e.g., the DLA34 network\cite{dla}), denoted by $\mathbf{F}_I$ for computing deep features from the raw image $I$,  and the regression heads for inferring the 3D bounding box parameters using the computed feature map.  
We generate bottom-up proposals using one off-the-shelf backbone monocular 3D object detector which is first trained following its own training receipts. Without loss of generality, we consider one of its detected 3D bounding boxes indexed by $i$ in an image $I$, 
\begin{equation}
    \mathbb{B}_i = (\ell_i, \mathbf{P}_i, \mathbf{D}_i, \alpha_i), \label{eq:proposal}
\end{equation}

As shown in Table~\ref{Tab:upper_bound}, the top-down sampling with a bottom-up anchor proposal is a straightforward process. We start by generating a 2D local grid in the $X$-$Z$ plane (i.e. in the bird's-eye-view, BEV), centered at the position $(X,Z)$ of the anchor box $\mathbb{B}$. To accomplish this, we specify a search range and a stride. Based on the trade-off between the empirical upper bound and the computing overhead and the mean IoU among the proposals, we adopt a conservative strategy using the search range $\pm1.5$ meters and the stride $0.75$, which will generate 25 proposals per bottom-up anchor (inclusive). We then place the anchor box $\mathbb{B}$ at the 25 grid points, with only the position $\mathbf{P}$ updated. To simplify the notation, we do not differentiate between the bottom-up anchor and the top-down sampled proposal, and we generally index them using $i$, unless otherwise stated.

% \begin{figure}[t!]
%     \centering
%     \includegraphics[width=0.48\textwidth]{figure/monoxiver_proposal_generation.png}
%     \caption{\textcolor{blue}{We generate a 2D grid for top-down 3D proposals based on the predicted bottom-up anchors in the X-Z plane. Then the top-down 3D proposals will be verified and rectified by our proposed MonoXiver.}}
%     \label{fig:topdown}
% \end{figure}

\subsection{The 3D-to-2D Proposal Verification}
The proposal verification will rely essentially on the information from the input 2D images in purely monocular 3D object detection. Due to the aforementioned unique challenges in the 3D proposal space, we utilize Perceiver I/O~\cite{perceiverio} to design an expressive proposal representation learning scheme.

\subsubsection{Proposal Representation} \label{sec:proposal_repr}
As illustrated in Fig.~\ref{fig:monoxiver}, we utilize three types of features in encoding a 3D bounding box proposal $\mathbb{B}$,
\begin{itemize} [leftmargin=*]
\itemsep 0em 
    \item \textit{The 3D Shape Features} represented by $(\mathbf{P}, \mathbf{D}, \alpha)$. It is a 7-dim vector. It encodes where the proposal is in the 3D space (the camera coordinate system), as well as its 3D occupancy. The position $\mathbf{P}$ is empirically normalized by the typical detection range (e.g., $X$, $Y$, $Z$ are normalized by 50, 2 and 80 meters respectively). % We have i) the 3D bounding box itself having 7 features; ii) the projected 2D bounding box in the image plane consisting of 4 features $(x, y, w, h)$, the center position and the sizes; iii) the projected 9 keypoints (8 corner points and the 3D center) having 18 features; The 2D geometric features are normalized by the image dimensions. In total, we have a 29-dim geometric feature vector. 
    \item \textit{The 2D Projection Points} from the 8 corners and the center of the 3D bounding box (with known camera intrinsic matrix). It is an 18-dim vector in the image coordinate system, and normalized by the image size. This 9-point projection provides the finegrained placement of a proposal in the 2D space, that is the geometric bond between a proposal in the 3D space and its 2D placement. 
    \item \textit{The 2D Bounding Box} projected from the proposal and truncated by the image boundary. We use the left-top and right-bottom two points to encode a 2D bounding box. It gives a 4-dim vector normalized by the image size. For a proposal whose 2D projection is entirely inside the image plane, the two points of the 2D bounding box will be just redundant with respect to the 9 project points. Otherwise, the truncation points on the image boundary will facilitate the learning to be truncation aware.     
    \item \textit{The Appearance Features} of the 9 projection points. We directly extract features from the final layer of the feature backbone $\mathbf{F}_I$. If a projection point is out of the image plane, we encode it using an all-zero vector. 
    \item \textit{The RoI Appearance Features} of the 2D bounding box. We use the RoIAlign features~\cite{maskrcnn} extracted from the final layer of the feature backbone $\mathbf{F}_I$. Since the 2D IoU overlapping between proposals is high on average (e.g., 0.612 in Table~\ref{Tab:upper_bound}), we use a finer grid, $14\times 14$, in computing the RoIAlign features, such that the extracted appearance features contain ``sufficient" details. 
\end{itemize}

\subsubsection{Proposal Embedding}

Let $N$ be the total number of proposals in an image $I$. With the above representation scheme, denote by $f^{geo}_{N\times 29}$ the geometric parameter matrix from the first three items above, and by $f^{pt}_{N\times 9\times C}$ the projection point appearance feature matrix (where $C$ is the number of channels of the final layer of the feature backbone, e.g., $C=64$), and by $f^{roi}_{N\times (14\times 14) \times C}$ the RoI feature matrix. Before applying the Perceiver I/O model, we embed the three to form ``tokens" using multi-layer perceptron (MLP). 

For the geometric parameters $f^{geo}_{N\times 29}$, we have,
\begin{equation}
    f^{geo}_{N\times 29}\xrightarrow[]{\text{MLP}}z_{N\times (g\times d)}\xrightarrow[]{\text{Rearrange}} \mathbf{f}^{geo}_{g\times N\times d},  
\end{equation}
where $d$ is the hidden dimension (e.g., $d=256$). $g$ is the group number, similar in spirit to the multi-head setup in the self-attention. We want to encode the geometric parameters in different latent spaces to account for the large variations in the proposal space (e.g., $g=4$).

For the project point features $f^{pt}_{N\times 9\times C}$, we have, 
\begin{equation}
    f^{pt}_{N\times 9\times C}\xrightarrow[]{\text{MLP}}z_{N\times 9\times d}\xrightarrow[]{\text{Rearrange}} \mathbf{f}^{pt}_{9\times N\times d}, 
\end{equation}
where we reuse $z$ to denote the latent features for simplicity and without confusion. 

For the RoI features $f^{roi}_{N\times (14\times 14) \times C}$, we have, 
\begin{equation}
    f^{roi}_{N\times (14\times 14) \times C}\xrightarrow[]{\text{MLP}}z_{N\times  d}\xrightarrow[]{\text{Rearrange}} \mathbf{f}^{roi}_{1\times N\times d}.  
\end{equation}

\subsubsection{Intra-Proposal Attention via Perceiver}

With the above proposal embedding, before the proposal verification, our goal is to compute a final latent feature representation in the $d$-dim space for each proposal by fusing information from the geometric encoding, projection-point-based appearance encoding and RoI-based appearance encoding (i.e., the intra-proposal attention), such that we can address the aforementioned unique challenges in the proposal space. We leverage the Perceiver I/O model for this intra-proposal attention. 

The Perceiver first fuses the projection-point appearance encoding $\mathbf{f}^{pt}_{9\times N\times d}$ and the geometric encoding $\mathbf{f}^{geo}_{g\times N\times d}$. As shown in Fig.~\ref{fig:monoxiver}, it consists of a cross-attention by treating the former as Query (consisting of 9 projection-point tokens) and computing Key and Value from the latter (consisting of $g$ geometric tokens), followed by a self-attention module. We have, 
\begin{equation}
    (\mathbf{f}^{pt}_{9\times N\times d},\mathbf{f}^{geo}_{g\times N\times d})\xrightarrow[]{\text{Cross-Attn}}\cdot \xrightarrow[]{\text{Self-Attn}} \mathbf{F}^{geo\text{-}pt}_{9\times N\times d},
\end{equation}
which results in geometry-aware projection-point encoding. 

Next, the Perceiver fuses the RoI appearance encoding $\mathbf{f}^{roi}_{1\times N\times d}$ as the 1-token Query with the 9-token geometry-aware projection-point encoding using a cross attention module, 
\begin{equation}
    (\mathbf{f}^{roi}_{1\times N\times d},\mathbf{F}^{geo\text{-}pt}_{9\times N\times d})\xrightarrow[]{\text{Cross-Attn}} \mathbf{F}^{geo\text{-}pt\text{-}roi}_{1\times N\times d},
\end{equation}
which results in the $d$-dim latent features for each proposal, which fuse the three types of information sources: geometry, point-level appearance and region-level appearance.

In the above, the cross-attention and self-attention modules are based on the standard formulation~\cite{attention-is-all-you-need}.

\subsubsection{Inter-Proposal Attention}

With all the above process, each proposal is still encoded individually with the hope of fusing geometry, point-level appearance and region-level appearance information in a semantically meaningful way via the Perceiver I/O model. With the compact latent vector computed for each proposal in $\mathbf{F}^{geo\text{-}pt\text{-}roi}_{1\times N\times d}$, their interactions need to be taken into account in order to resolve their ``crowding" issue at the underlying scene level. 

To that end, we treat each proposal as a ``token" and apply a standard self-attention module to capture the inter-proposal attention, 
\begin{equation}
    \mathbf{F}^{geo\text{-}pt\text{-}roi}_{1\times N\times d}\xrightarrow[]{\text{Rearrange}} \mathbf{F}^{geo\text{-}pt\text{-}roi}_{N\times d} \xrightarrow[]{\text{Self-Attn}}  \mathbb{F}_{N\times d}. 
\end{equation}

\subsubsection{Proposal Verification}
This verification is posed as a regression problem , 
\begin{equation}
    \mathbb{F}_{N\times d} \xrightarrow[]{\text{MLPs}} (\mathbf{y}_{N\times L}, \Delta\mathbf{P}_N, \Delta\mathbf{D}_N), \label{eq:regression}
\end{equation}
where $L$ is the total number of categories (e.g., $L=3$ in the KITTI dataset), and $\mathbf{y}_{N\times L}$ is the classification scores (logits). $\Delta\mathbf{P}_N$ and $\Delta\mathbf{D}_N$ are the position and dimension residuals. 

Consider a proposal $\mathbb{B}_i$ (Eqn.~\ref{eq:proposal}), it is verified via, 
\begin{align}
    \hat{\ell}_i &= \arg\max_{l=1,\cdots, L} \mathbf{y}_{i,l}, \\
    \hat{\mathbf{P}}_i &= \mathbf{P}_i + \Delta\mathbf{P}_i, \\
    \hat{\mathbf{D}}_i & = \mathbf{D}_i + \Delta\mathbf{D}_i,
\end{align}
where a proper unnormalization step will be done for $\hat{\mathbf{P}}_i$ to counter the normalization step used in the proposal representation (Sec.~\ref{sec:proposal_repr}). Based on the score $\mathbf{y}_{i,\hat{\ell}_i}$, we can keep top-$k$ proposals for each anchor position.

\subsection{Details of Training and Testing}
For simplicity, we evaluate the proposed MonoXiver in a two-stage setting where the backbone monocular 3D object detector is first trained and kept frozen. Then, the MonoXiver component is trained end-to-end. One of the reasons for this choice is that off-the-shelf backbone monocular 3D object detectors often involve multiple loss functions, and joint training with MonoXiver would require sophisticated tuning of the trade-off parameters for different loss terms. We leave the joint end-to-end training or iterative training, as done in the early version of Faster RCNN~\cite{fasterrcnn} between the region proposal network and the region classification head network, for future work. Additionally, the separate two-stage training may also have advantages in leveraging the set of bottom-up anchor proposals merged from multiple backbone monocular 3D detectors, which we also leave for future work.

\subsubsection{The Set Prediction Formulation in Training}
Based on the separate two-stage setting, the proposed MonoXiver is trained with a fixed set of 3D bounding box proposals and a fixed set of ground-truth 3D bounding boxes. The ground-truth assignment for proposals is needed in training. Two straightforward methods are: using the maximum 3D IoU based assignment, or using the maximum 2D IoU assignment (after projected to the image plane). Due to the ``crowding" issue in the proposal space, we observe that both of them do not work during our development of the MonoXiver since they will create difficult ``decision boundaries" for the MonoXiver to learn or fit.

Since we have the two fixed set as input, we resort to the set prediction formulation  used in the DETR framework~\cite{detr}. Denote by $\Omega_{\mathbb{B}}=\{\mathbb{B}_i\}_{i=1}^N$ the set of generated proposals, by $\Omega_{\mathbb{B}^*}=\{\mathbb{B}^*_j\}_{j=1}^M\cup \{\emptyset_j\}_{j=M+1}^N$ the set of ground-truth 3D bounding boxes padded with $N-M$ dummy $\emptyset$ elements. Given a permutation of $N$
elements, denoted by $\sigma$ which assigns the $i$-th element in the ground-truth set to the $\sigma(i)$-the element in the proposal set,  the loss for the one-to-one bipartite matching between $\mathbb{B}^*_i$ and $\mathbb{B}_{\sigma(i)}$ is defined by, 
\begin{align}
    \mathcal{L}_{match}(\mathbb{B}^*_i, \mathbb{B}_{\sigma(i)}) & = \mathbf{1}_{\ell_i^*\neq \emptyset}\cdot  ( -\lambda_1\cdot \hat{p}_{\sigma(i)}(\ell^*_i) + \\
    \nonumber & \lambda_2\cdot \mathcal{L}_{bbox}^{2D} + \lambda_3\cdot \mathcal{L}_{iou}^{2D} + \lambda_4\cdot \mathcal{L}_{iou}^{3D}),
\end{align}
where $\hat{p}_{\sigma(i)}(\ell^*_i)$ is computed using $\mathbf{y}_{\sigma(i)}$ (Eqn.~\ref{eq:regression}) via Softmax.  $\mathcal{L}_{bbox}^{2D}$ represents the normalized L1 2D bounding box, and $\mathcal{L}_{iou}^{2D}$ and $\mathcal{L}_{iou}^{3D}$ the IoU loss in 2D and 3D respectively. $\lambda_1$ to $\lambda_4$ are the trade-off parameters.

% \textbf{Ground Truth Assignment for Proposals in Training.} In the line of Region-based CNN for 2D object detection \cite{fastrcnn, fasterrcnn, maskrcnn}, a positive prediction is assigned to a ground truth based on the maximum intersection-over-union (IoU) criterion. However, the maximum 2D IoU assignment policy might fail in selecting a better bounding box in 3D space because projected 2D IoU in image is not aligned with 3D IoU in camera coordinate. Therefore, the trade-off between focusing 2D or 3D need to be made. We extend the set assignment method proposed by \cite{detr} considering the contribution of both 2D and 3D space. The total cost is defined as:

% \begin{equation}
%     \mathcal{C} = \lambda_1 \cdot \mathcal{C}_{cls} + \lambda_2 \cdot \mathcal{C}_{2DL1} + \lambda_3 \cdot \mathcal{C}_{2DIoU}  + \lambda_4 \cdot \mathcal{C}_{3DIoU} \label{eq:cost} 
% \end{equation}

%  where we use the focal loss cost for $\mathcal{C}_{cls}$, normalized L1 2D bounding box cost for $\mathcal{C}_{2DL1}$, 2D IoU cost and 3D IoU cost. The empirical trade-off weight is set to be 2, 5, 2, 8.

\subsubsection{Loss Functions for Proposal Verification} 
For the three outputs in Eqn.~\ref{eq:regression}, we have three loss terms as follows, 
\begin{equation}
    \mathcal{L} = \mathcal{L}_{cls} + \mathcal{L}_{size} + \lambda \cdot \mathcal{L}_{loc} \label{eq:loss} 
\end{equation}
where $\mathcal{L}_{cls}$ is the Focal Loss \cite{focal} with the $\alpha$ value of 0.5 to balance the number of positive and negative samples. The Focal loss is used due to the imbalance introduced by the explicit top-down proposal generation. $\mathcal{L}_{size}$ are  $\mathcal{L}_{loc}$ are the loss functions for the 3D size and the 3D center respectively. $\lambda$ is an trade-off hyper-parameter. We use $\ell_1$ loss for $\mathcal{L}_{size}$ and $\mathcal{L}_{loc}$. We set $\lambda=5$ to induce the model to focus more on the 3D localization refinement task.

\subsubsection{Training.} MonoXiver is trained using a total of 8 GPUs, with a batch size of 64 for 24 epochs on KITTI (12 epochs on Waymo) using the AdamW optimizer. The optimizer is set with $(\beta_1, \beta_2)=(0.95, 0.99)$ and weight decay of 0.0001, excluding feature normalization layers and bias parameters. The initial learning rate is set to $2.25e-5$, and it is reduced by a factor of 10 at the 16th and 22nd epoch. Notably, the entire second-stage training process takes merely 1.5 hours on KITTI \cite{kitti} using a single Nvidia RTX5000 GPU, indicating the low computational overhead of our proposed MonoXiver method. The training time on Waymo~\cite{waymo} varies depending on the training data and training recipes of off-the-shelf monocular 3D detectors. Overall, the second-stage training on Waymo is much faster compared to training the backbone monocular 3D object detector, leading to about a 3/4 reduction in training time.

\subsubsection{Testing} 

During testing, we keep at most top-3 predictions per initial anchor proposal with a score threshold of 0.03 (the score is predicted from the proposed MonoXiver module).
The final prediction score is the product of the classification probability predicted from the backbone monocular 3D object detector and the score computed by the MonoXiver module. 

\begin{table*}[t]
\begin{center}
\vspace{2mm}
\resizebox{0.75\linewidth}{!}{
    \begin{tabular}{l|c|c|ccc|ccc}
    \toprule
    \multirow{2}{*}{Methods} & \multirow{2}{*}{Venues} & \multirow{2}{*}{Extra Info.} &
    \multicolumn{3}{c|}{\textit{$AP_{BEV|R40}\uparrow$}} & \multicolumn{3}{c}{\textit{$AP_{3D|R40}\uparrow$}} \\
     & &  & Easy & Mod. & Hard  & Easy & \textbf{Mod.} & Hard \\
    \midrule
    
    Kinematic3D \cite{kinematic3d} & \textit{ECCV20} & Temporal & 26.69 & 17.52 & 13.10 & 19.07 & 12.72 & 9.17 \\
    
    AutoShape \cite{autoshape} & \textit{ICCV21} & CAD & 30.66 & 20.08 & 15.95 & 22.47 & 14.17 & 11.36 \\
    
    DCD \cite{dcd} & \textit{ECCV22} & CAD & 32.55 & 21.50 & 18.25 & 23.81 & 15.90 & 13.21 \\
    
    % \midrule

    % CaDDN \cite{caddn} & \textit{CVPR21} & \multirow{5}{*}{Lidar} & 27.94 & 18.91 & 17.19 & 19.17 & 13.41 & 11.46 \\
    
    % MonoDTR \cite{monodtr} & \textit{CVPR22} & & 28.59 & 20.38 & 17.14 & 21.99 & 15.39 & 12.73 \\
    
    MonoDistill \cite{monodistill} & \textit{ICLR22} & LiDAR & 31.87 & 22.59 & 19.72 & 22.97 & 16.03 & 13.60 \\
    
    % MonoJSG \cite{monojsg} & \textit{CVPR22} & & 32.59 & 21.26 & 18.18 & 24.69 & 16.14 & 13.64 \\
    
    DID-M3D \cite{did} & \textit{ECCV22} & LiDAR & \underline{32.95} & 22.76 & 19.83 & \underline{24.40} & 16.29 & 13.75 \\

    % \midrule

    % PatchNet \cite{patchnet} & \textit{ECCV20} & \multirow{5}{*}{Depth} & 22.97 & 16.86 & 14.97 & 15.68 & 11.12 & 10.17\\
    
    % Demystifying \cite{demystifying} & \textit{ICCV21} & & 29.97 & 17.70 & 15.04 & 22.40 & 12.53 & 10.64 \\
    
    % DDMP-3D \cite{ddmp3d} & \textit{CVPR21} & & 28.08 & 17.89 & 13.44 & 19.71 & 12.78 & 9.80\\
    
    % DFRNet \cite{dfrnet} & \textit{ICCV21} & Depth & 28.17 & 19.17 & 14.84 & 19.40 & 13.63 & 10.35 \\
    
    DD3D \cite{dd3d} & \textit{ICCV21} & Depth & 30.98 & 22.56 & 20.03 & 23.22 & 16.34 & 14.20 \\

    % \midrule

    DFM \cite{dfm} & \textit{ECCV22} & Temporal + LiDAR & 31.71 & 22.89 & 19.97 & 22.94 & 16.82 & 14.65 \\
    
    Pseudo-Stereo \cite{pseudostereo} & \textit{CVPR22} & Depth + LiDAR & {32.84} & \underline{23.67} & \underline{20.64} & {23.74} & \underline{17.74} & \underline{15.14} \\ 
    
    \midrule

    MonoFlex \cite{monoflex} & \textit{CVPR21} & \multirow{5}{*}{None} & 28.23 & 19.75 & 16.89 & 19.94 & 13.89 & 12.07\\
    
    GUPNet \cite{gupnet} & \textit{ICCV21} & & 30.29 & 21.19 & 18.20 & 20.11 & 14.20 & 11.77 \\
    
    DEVIANT \cite{deviant} & \textit{ECCV22} & & 29.65 & 20.44 & 17.43 & 21.88& 14.46 & 11.89 \\
    
    Homography \cite{homography} & \textit{CVPR22} & & 29.60 & 20.68 & 17.81 & 21.75 & 14.94 & 13.07 \\
    
    DimEmbedding \cite{dimembed} & \textit{CVPR22} & & {32.82} & 21.98 & 18.70 & {23.62} & 16.10 & 13.41 \\
    
    % MonoDDE \cite{monodde} & \textit{CVPR22} & & \underline{33.58} & 23.46 & 20.37 & \underline{24.93} & 17.14 & 15.10 \\

    \midrule
    
    MonoCon \cite{monocon} & \textit{AAAI22} &  \multirow{2}{*}{None} & 31.12 & {22.10} & {19.00} & 22.50 & {16.46} & {13.95}\\

    \textbf{Our MonoXiver} + MonoCon & - & & \textbf{34.14} & \textbf{25.37} & \textbf{22.20} & \textbf{25.24} & \textbf{19.04} & \textbf{16.39} \\

    \bottomrule
    \end{tabular}}
\end{center}
\vspace{-4mm}
\caption{\textbf{Comparisons with state-of-the-art methods on the Car category on the KITTI \textit{test} set.} According to the KITTI protocol, methods are ranked based on their performance under the moderate difficulty setting. We highlight the best results in \textbf{bold} and the second-best results in \underline{underline}. } \label{Tab:comparison_test} 
\vspace{-2mm}
\end{table*}

\begin{table}[h]
\begin{center}
\resizebox{0.9\linewidth}{!}{
\begin{tabular}{l|ccc|c}
\toprule
\multirow{2}{*}{Methods} & \multicolumn{3}{c|}{AP$_{3D}$$|_{R_{40}}\uparrow$} & Relative\\
 & Easy & Moderate & Hard & Improvement\\
\midrule
 
SMOKE \cite{smoke} & 10.43 & 7.09 & 5.57 & \multirow{3}{*}{11\%-33\%}\\

SMOKE + Ours & 11.58 & 9.40 & 7.75 \\

\textit{Improvement} & +1.15 & +2.31 & +2.18 \\

\midrule

MonoDLE \cite{monodle} & 17.94 & 13.72 & 12.10 & \multirow{3}{*}{18\%-22\%}\\

MonoDLE + Ours & 21.15 & 16.19 & 14.75  \\

\textit{Improvement} & +3.21 & +2.47 & +2.65 \\

\midrule

MonoCon \cite{monocon} & 26.33 & 19.01 & 15.98 & \multirow{3}{*}{16\%-20\%}\\

MonoCon + Ours & 30.48 & 22.40 & 19.13 \\

\textit{Improvement} & +4.15 & +3.39 & +3.15 \\

\bottomrule
\end{tabular}
}
\end{center}
\vspace{-4mm}
\caption{\textbf{The effectiveness of MonoXiver based on different methods on the KITTI validation dataset.} In order to demonstrate the ability of the proposed MonoXiver to generalize across different detectors, we utilized three distinct base detectors with varying levels of detection accuracy. We observe that MonoXiver consistently yields significant improvements across all three base detectors.}
\vspace{-2mm}
\label{Tab:relative_improvement_val}
\end{table}

\begin{table}[h]
\begin{center}
\resizebox{0.9\linewidth}{!}{
\begin{tabular}{l|cc|cc}
\toprule
\multirow{2}{*}{Methods} & \multicolumn{2}{c|}{Level\_1} & \multicolumn{2}{c}{Level\_2} \\
 & AP$_{3D}$$\uparrow$ & APH$_{3D}$$\uparrow$ & AP$_{3D}$$\uparrow$ & APH$_{3D}$$\uparrow$ \\
\midrule

PatchNet \cite{patchnet} & 2.92 & 2.74 & 2.42 & 2.28 \\

PCT \cite{pct} & 4.20 & 4.15 & 4.03 & 4.15 \\

\midrule

GUPNet \cite{gupnet} & 10.02 & 9.94 & 9.39 & 9.31 \\

GUPNet + ours & 11.47 & 11.35 & 10.67 & 10.56  \\

\textit{Improvement} & +1.45 & +1.41 & +1.28 & +1.25 \\

\midrule

DEVIANT \cite{deviant} & 10.98 & 10.89 & 10.29 & 10.20 \\

DEVIANT + ours &  11.88 & 11.75 & 11.06 & 10.93 \\

\textit{Improvement} & +0.90 & +0.86 & +0.77 & +0.73 \\

\bottomrule
\end{tabular}
}
\end{center}
\vspace{-4mm}
\caption{\textbf{The effectiveness of MonoXiver based on different methods on Waymo validation dataset.} In order to showcase the generalization capabilities of the proposed MonoXiver on large-scale datasets, we conducted testing on the challenging Waymo \cite{waymo} validation set, utilizing two SOTA methods. We observed consistent improvements in performance.}
\vspace{-2mm}
\label{Tab:relative_improvement_waymo}
\end{table}

%%%%%%%%%%%%%%%%
\begin{table}[ht]
\begin{center}
\resizebox{0.9\linewidth}{!}{
    \begin{tabular}{c|c|c|c}
    \toprule
    Methods & FLOPs & Latency (ms) & FPS \\
    \midrule
    
    SMOKE \cite{smoke}/ + ours & 42.82/48.19 & 23/31 & 43/32 \\
    
    \midrule

    MonoDLE \cite{monodle}/ + ours & 77.44/82.80 & 22/31 & 45/32 \\

    % {MonoDLE + Ours} & 79.04 & 25.00M \\
    
    \midrule

    MonoCon \cite{monocon}/ + ours & 56.22/61.50 & 18/25 & 55/40 \\
    
    \bottomrule
    \end{tabular}}
\end{center}
\vspace{-4mm}
\caption{\small \textbf{Computation overhead analysis}. The GFLOPs is computed based on an input size of $(384, 1248)$, while the latency and FPS are evaluated on a NVIDIA RTX5000 GPU.}
\vspace{-2mm}
\label{Tab:flops}
\end{table}

\section{Experiments}

We evaluate our MonoXiver on the well-established KITTI~\cite{kitti} dataset and the large-scale Waymo~\cite{waymo} dataset. We first show that our improved baseline detector outperforms previous SOTA methods on the KITTI benchmark by a large margin. Then we show that we consistently improved pretrained monocular 3D object detectors with limited computation overhead on both datasets. We further analyze the contribution of each part of our MonoXiver in the ablation studies. 

\subsection{Setup}

\textbf{Dataset.} 
\textbf{The KITTI dataset}~\cite{kitti} consists of 7,481 training images and 7,518 testing images. 
In KITTI's experiments, we split the training data into a training subset with 3,712 images and a validation subset with 3,769 images following \cite{mono3d, chen2, multiview}. We conduct ablation studies on the defined split and report the test set results evaluated by the official KITTI benchmark. \textbf{The Waymo dataset}~\cite{waymo} contains 52,386 training and 39,848 validation images from the front camera. In Waymo experiments, we use a subset of its training set by sampling every third frame from the training sequences following \cite{deviant, caddn}.

\textbf{Evaluation Metrics.} The KITTI vehicle benchmark evaluates detection results by the 40-point interpolated average precision ($AP_{R40}$) of 3D bounding boxes in 3D space ($AP_{3D|R40}$) and bird eye's view ($AP_{BEV|R40}$) at $IoU_{3D}\geq0.7$. The prediction results are evaluated based on three difficulty settings, {\tt easy, moderate} and {\tt hard}, according to the 2D box height, occlusion and truncation levels of objects. \textbf{The vehicle detection results on Waymo} are evaluated on two levels of difficulty including {\tt Level\_1} and {\tt Level\_2} at $IoU_{3D}\geq0.5$. The level is assgined based on the number of LiDAR points included in each 3D box. Besides the $AP_{3D}$ metric, Waymo uses the $APH_{3D}$ metric to incorporate heading information in $AP_{3D}$.

\subsection{Experimental Results}

%%%%%%%%%%%%%%%

\noindent \textbf{Comparison with SOTA methods on the KITTI dataset.} We present the results of our proposed MonoXiver method on the challenging KITTI vehicle benchmark in Table~\ref{Tab:comparison_test}. Notably, our method achieves the best performance across different evaluation metrics while only using image-level information, surpassing previous state-of-the-art methods by a large margin. Specifically, we observe a significant and consistent improvement from \textbf{2.44} AP ({\tt hard} settings) to \textbf{2.74} AP ({\tt easy} settings) \textbf{absolute increase} in AP$_{3D}$ for the boosted MonoCon approach compared to the vanilla MonoCon approach. These consistent improvements demonstrate the effectiveness of our method. More qualitative and quantitative results are provided in the \textcolor{blue}{Supplementary Materials}.

\vspace{0.2em}
\noindent \textbf{Generalization abilities across backbone monocular 3D object detectors and datasets.} We evaluate our method with various state-of-the-art backbone monocular 3D object detection methods on the KITTI and the Waymo validation set, reported in Table~\ref{Tab:relative_improvement_val} and Tabel~\ref{Tab:relative_improvement_waymo} respectively. 
Monoxiver consistently demonstrates significant improvements across different methods and different difficult levels. Specifically, on the well-established KITTI dataset, our MonoXiver is able to enhance various backbone detectors with varying levels of detection accuracy by up to 33\% relative improvement. On the challenging large-scale Waymo dataset, our MonoXiver consistently improves the performance of strong baseline methods GUPNet and DEVIANT by up to 1.45 AP$_{3D}$ on {\tt Level\_1} and 1.28 AP$_{3D}$ on {\tt Level\_2}. These results validate the effectiveness and robustness of our approach.

\noindent \textbf{Computation overhead.} As shown in Table~\ref{Tab:flops}, although our MonoXiver causes about an average of 8 ms overhead compared with baseline detectors, it still achieves real-time detection. We note that its inference speed could be improved by optimizing the detailed implementation (e.g., we still have for-loops in our current code, which can be easily paralleled for better efficiency). 

\begin{figure}[h]
    \centering
    % \vspace{-2mm}
    \includegraphics[width=0.95\linewidth]{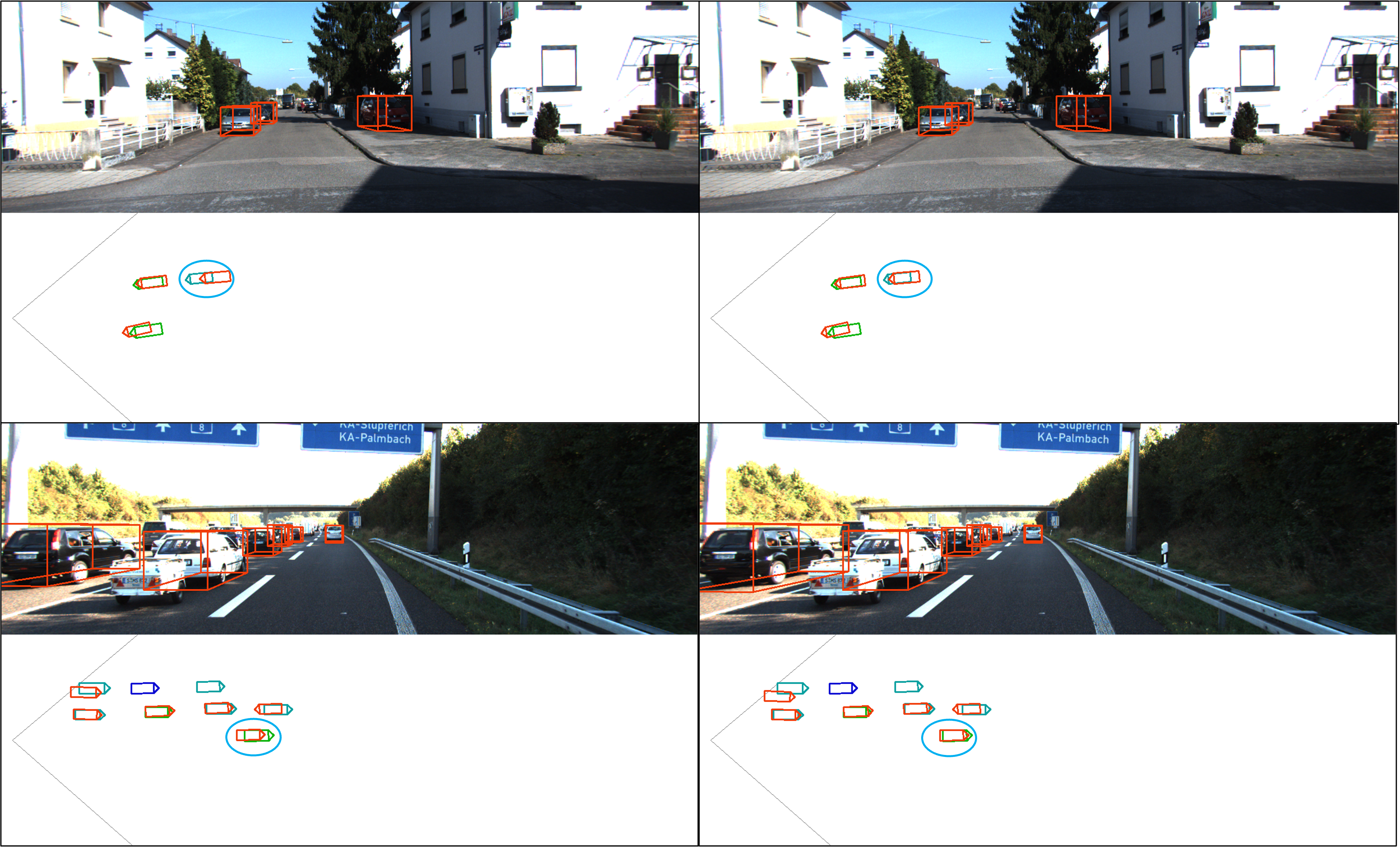}
    \includegraphics[width=0.94\linewidth]{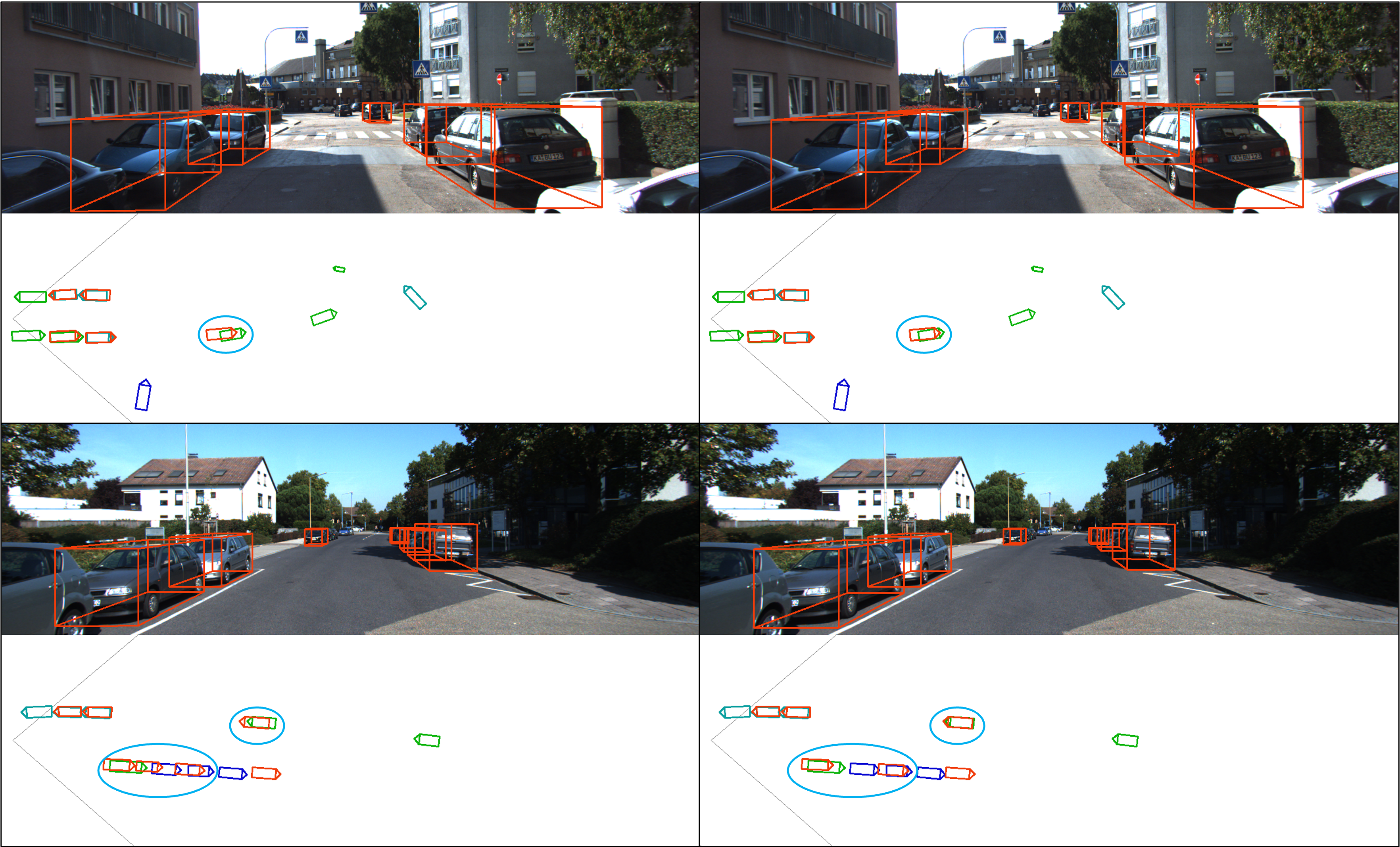}
    % \vspace{-2mm}
    \caption{\textbf{Qualitative comparison of our MonoXiver with MonoCon on KITTI \textit{validation} set \cite{mono3d}.} The left column is MonoCon's prediction result; the right column is our MonoXiver's prediction result. The ground truth is shown in \textcolor{green}{green} and \textcolor{blue}{blue}. The prediction result is shown in \textcolor{red}{red}.}
    \label{fig:qual_val}
    % \vspace{-4mm}
\end{figure}

\noindent \textbf{Qualitative Comparison.} We show the visualization comparisons with MonoCon in Fig.~\ref{fig:qual_val}. It shows that MonoXiver achieves better 3D box center localization.

\subsection{Ablation Studies} \label{sec:ablation}
In this section, we report ablation studies of MonoXiver structure and different bounding box branches on the KITTI validation set. More ablation studies are provided in the \textcolor{blue}{Supplementary Materials}.

\begin{table} [h!]
\begin{center}
% \vspace{-2mm}
\resizebox{0.9\linewidth}{!}{
    \begin{tabular}{c|c|c|c|c}
    \toprule
    & Appearance & Geometry & Perciver & Easy/Mod./Hard \\
    \midrule
    
    MonoCon \cite{monocon} & - & - & - & 26.33/19.01/15.98\\ 
    
    a. & \checkmark & - & - & 29.30/21.04/18.22 \\
    
    b. & \checkmark & \checkmark & - & 29.33/21.67/18.46 \\
    
    c. & \checkmark & \checkmark & \checkmark & 30.48/22.40/19.13 \\

    \bottomrule
    \end{tabular}
}
\end{center}
\vspace{-4mm}
\caption{Ablation studies of MonoXiver structure on KITTI validation set.} \label{Tab:ablation}
\vspace{-2mm}
\end{table}

\begin{table} [h!]
\begin{center}
\resizebox{0.9\linewidth}{!}{
    \begin{tabular}{c|c|c|c|ccc}
    \toprule
    & Rescore & Res$_{Loc}$ & Res$_{Dim}$ & Easy & Mod. & Hard \\
    \midrule
    
    MonoCon \cite{monocon} & - & - & - & 26.33 & 19.01 & 15.98 \\
    
    a. & \checkmark & - & - & 29.66 & 21.41 & 18.38 \\
    
    b. & \checkmark & \checkmark & - & 30.32 & 22.30 & 19.04 \\
    
    c. & \checkmark & \checkmark & \checkmark & 30.48 & 22.40 & 19.13 \\
    
    \bottomrule
    \end{tabular}
    }
\end{center}
\vspace{-4mm}
\caption{Ablation studies of the effect of different branches on KITTI validation set} \label{Tab:ablation_branch}
\vspace{-2mm}
% \vspace{-7mm}
\end{table}

\noindent \textbf{Effectiveness of MonoXiver structures.} In this study, we explore the importance of RoI feature (Table~\ref{Tab:ablation} a.), the context information provided by geometric embeddinWgs (Table~\ref{Tab:ablation} b.), and our complete MonoXiver model (Table~\ref{Tab:ablation} c.). In experiments a. and b., we replace the Perciver with a MLP layer to fuse the appearance feature and geometry feature to keep similar parameter size. 

Results show that the appearance feature plays the most important role in performance improvement. When using the appearance feature only, the Moderate AP is improved by 2.39. The performance is further improved by 0.6 AP and 0.7 AP after fusing with geometric embeddings and Perciver. It shows that Perceiver effectively fuses information between appearance and geometric embeddings.

\noindent \textbf{Effectiveness of Different Head Branches.} In Table~\ref{Tab:ablation_branch}, we present the impact of different branches on the detection performance of our MonoXiver. Since our method generates 25 proposals per each initial proposal, the rescoring head branch plays a crucial role in filtering out false detections, leading to significant improvements by up to 3.3 AP. The localization residual branch and the dimension residual branch further enhance the performance by about 1 AP, and we employ all branches to achieve a new state-of-the-art result.

\vspace{-2mm}
\section{Conclusion}
\vspace{-2mm}
This paper explores the task of monocular 3D object detection. It begins with the strong observation that significant improvements in detection performance can be achieved by a local grid search based on initially detected 3D bounding boxes. This leads to a novel denoising-based approach called MonoXiver. The approach utilizes the Perceiver I/O model to produce a unified feature embedding, leverages the self-attention mechanism for proposal verification, and ultimately delivers high-quality 3D box predictions. MonoXiver can serve as a lightweight second-stage processing module to significantly improve the accuracy of detection results while only requiring 1.5 hours of training on a single GPU card for the KITTI dataset. Comprehensive evaluation experiments on the well-established KITTI benchmark and the large-scale challenging Waymo dataset demonstrate the effectiveness of our design, with a new state-of-the-art achieved.
Furthermore, our strong observation should encourage the community to further study refinement modules for 3D object detection.

\section*{Acknowledgements} \vspace{-2mm}
{This research is partly supported by  ARO Grant W911NF1810295, NSF IIS-1909644, ARO Grant W911NF2210010, NSF IIS-1822477, NSF CMMI-2024688 and NSF IUSE-2013451. 
The views and conclusions contained herein are those of the authors and should not be interpreted as necessarily representing the official policies or endorsements, either expressed or implied, of the ARO, NSF, DHHS or the U.S. Government. The U.S. Government is authorized to reproduce and distribute reprints for Governmental purposes not withstanding any copyright annotation thereon. 
%The authors are grateful for the constructive comments by anonymous reviewers and area chairs. 
}

{\small
\bibliographystyle{ieee_fullname}
\bibliography{egbib}
}

\clearpage
% \newpage 
% \setcounter{section}{0}
% \renewcommand{\thesection}{\Alph{section}}
\appendix
%%%%%%%%% BODY TEXT - ENTER YOUR RESPONSE BELOW
\section*{Supplementary Materials Overview}
In this supplementary material, we provide more details on the following aspects that are not presented in the main paper due to space limit: 
\begin{itemize}
    \item \textit{Results on the KITTI Validation Set} are provided in Sec.~\ref{sec:val_results}.  
    \item \textit{Supplementary ablation studies} are provided in Sec.~\ref{sec:ablation_supp}. 
    \item \textit{Results and analysis on other category} are provided in Sec.~\ref{sec:other}.  
    \item \textit{Qualitative failure case study on the KITTI Validation Set} are provided in Sec.~\ref{sec:qual_kitti}.  
    \item \textit{Qualitative failure case study on the Waymo Validation Set} are provided in Sec.~\ref{sec:qual_waymo}.  
    \item \textit{Implementation details} are provided in Sec.~\ref{sec:network}.  
\end{itemize}

\section{Supplementary Results}

\subsection{Results on the KITTI Validation Set} \label{sec:val_results}

We report quantitative results of car category on the KITTI validation set as shown in Tab.~\ref{Tab:val}. 
It shows that our MonoXiver obtains significant improvements with different backbone detectors including SMOKE~\cite{smoke} and MonoCon~\cite{mono3d}.

\begin{table}[h]
\begin{center}
\resizebox{0.46\textwidth}{!}{
\begin{tabular}{p{3cm}|p{2cm}|ccc|ccc}
\toprule
\multirow{2}{*}{Methods} & \multirow{2}{*}{Extra} & \multicolumn{3}{c|}{\textit{$AP_{3D|R40}$}} & \multicolumn{3}{c}{\textit{$AP_{BEV|R40}$}}\\
 & & Easy & Mod. & Hard & Easy & Mod. & Hard \\
\midrule

Kinematic3D \cite{kinematic3d} & Temporal & 19.76 & 14.10 & 10.47 & 27.83 & 19.72 & 15.10 \\

\midrule

DFM \cite{dfm} & Temporal \& Lidar & \underline{29.27} & \underline{20.22} & \underline{17.46} & \underline{38.60} & \underline{27.13} & \underline{24.05} \\

\midrule

DID-M3D \cite{did} & \multirow{5}{*}{Lidar} & 22.98 & 16.12 & 14.03 & 31.10 & 22.76 & 19.50 \\

CaDDN \cite{caddn} &  & 23.57 & 16.31 & 13.84 & - & - & - \\

MonoJSG \cite{monojsg} &  & 26.40 & 18.30 & 15.40 & - & - & - \\

MonoDistill \cite{monodistill} & & 24.31 & 18.47 & 15.76 & 33.09 & {25.40} & {22.16} \\

MonoDTR \cite{monodtr} & & 24.52 & 18.57 & 15.51 & 33.33 & 25.35 & 21.68 \\

\midrule
    
MonoFlex \cite{monoflex} & \multirow{5}{*}{None} & 23.64 & 17.51 & 14.83 & - & - & - \\
    
GUPNet \cite{gupnet} & & 22.76 & 16.46 & 13.72 & 31.07 & 22.94 & 19.75 \\

DEVIANT \cite{deviant} & & 24.63 & 16.54 & 14.52 & 32.60 & 23.04 & 19.99 \\

Homography \cite{homography} & & 23.04 & 16.89 & 14.90 & 31.04 & 22.99 & 19.84 \\

\midrule

SMOKE \cite{smoke} & \multirow{4}{*}{None} & 10.43 & 7.09 & 5.57 & 17.62 & 12.02 & 10.07   \\

\textbf{Ours} + SMOKE & & 11.58 & 9.40 & 7.75 & 18.07 & 14.47 & 12.01 \\

MonoCon \cite{monocon} & & {26.33} & {19.01} & {15.98} & {34.65} & 25.39 & 21.93\\

\textbf{Ours} + MonoCon & & \textbf{30.48} & \textbf{22.40} & \textbf{19.13} & \textbf{38.77} & \textbf{28.67} & \textbf{24.89}\\

\bottomrule
\end{tabular}
}
\end{center}\vspace{-1mm}
\caption{Quantitative performance of the \textbf{Car} category on the KITTI \textit{validation} set. Method are ranked by moderate settings based on 3D detection performance following KITTI leaderboard within each group. We highlight the best results in \textbf{bold} and the second place in \underline{underline}.}
\label{Tab:val} \vspace{2mm}
\end{table}

\subsection{Supplementary Ablation Studies} \label{sec:ablation_supp}

\noindent \textbf{Choice of Top-down Generated Proposal Anchors:} We report the result of generating a different number of top-down proposals in Tab.~\ref{Tab:proposal}. When we reduce the number of proposals or the range of generated proposals (Tab.~\ref{Tab:proposal} b. v.s. Tab.~\ref{Tab:proposal} c. and Tab.~\ref{Tab:proposal} d.), the performance will drop a lot. This is because the detection performance is dependent on the quality of anchors (recall rate), which aligns with our empirical upper-bound analysis presented in Sec. 3 of the main paper well. When we increase the number of the top-down proposals (Tab.~\ref{Tab:proposal} a.), the performance also drops. The reason might be that the densely generated proposals will heavily overlap with each other in 2D feature maps (as discussed in the introduction of our main paper). This will make the model confused, and lead to the difficulty of optimization during training.

\begin{table} [h!]
\begin{center}

\resizebox{0.8\linewidth}{!}{
    \begin{tabular}{c|c|c|c|c}
    \toprule
    & Range & Stride & \#Bboxes & Easy/Mod./Hard \\
    \midrule
    
    MonoCon \cite{monocon} & - & - & - & 26.33/19.01/15.98\\ 
    
    a. & 1.5 & 0.5 & 49 & 29.34/21.18/17.82 \\
    
    b. & 1.5 & 0.75 & 25 & \textbf{30.48}/\textbf{22.40}/\textbf{19.13} \\
    
    c. & 1.5 & 1.5 & 9 & 27.32/19.72/16.67  \\
    
    d. & 1.0 & 0.5 & 25 & 28.80/20.97/17.61 \\

    \bottomrule
    \end{tabular}
}
\end{center}
\vspace{-3mm}
\caption{Ablation studies on the top-down proposal generation. Setting b. is used in our main experiments in the main paper.} \label{Tab:proposal}
% \vspace{-3mm}
\end{table}

\begin{table} [h!]
\begin{center}

\resizebox{0.8\linewidth}{!}{
    \begin{tabular}{c|c|c|c|c}
    \toprule
    & Cls & 2D Box & 3D Box & Easy/Mod./Hard \\
    \midrule
    
    MonoCon \cite{monocon} & - & - & - & 26.33/19.01/15.98\\ 
    
    a. & \checkmark & \checkmark & - & 22.54/15.73/12.99 \\
    
    b. & \checkmark & - & \checkmark &  30.09/22.08/18.92 \\
    
    c. & \checkmark & \checkmark & \checkmark & \textbf{30.48}/\textbf{22.40}/\textbf{19.13} \\
    
    d. & - & - & - & 10.18/8.41/7.23 \\

    \bottomrule
    \end{tabular}
}
\end{center}
\vspace{-3mm}
\caption{Ablation studies on the bounding box assigner. The setting c. is used in our main experiments in the main paper.} \label{Tab:assinger}
% \vspace{-6mm}
\end{table}

\noindent \textbf{Design of Ground Truth Assignment:} We use set-prediction formulation during training. The bipartite matching consists of four costs: 1) classification cost, 2) 2D bounding box $L_1$ cost, 3) 2D IoU cost, 4) 3D IoU cost. We report detailed ablations in Tab.~\ref{Tab:assinger}. 
Tab.~\ref{Tab:assinger} a. shows that the performance will drop a lot if we only use the 2D bounding boxes for assignment. This implies that the quality of the 2D box cannot ensure the prediction quality in 3D. Tab.~\ref{Tab:assinger} b. shows that only using the 3D box as an assignment basis will also lead to a performance drop compared with Tab.~\ref{Tab:assinger} c. This is because there are many cases in which predicted bottom-up proposals have no overlap with ground truth in 3D space. 
In these cases, the 2D box terms will serve as an auxiliary criterion during training to help the model select highly related 2D regions in the feature map to predict 3D boxes. We also try to use max IoU-based assignment criterion following Faster-RCNN \cite{fasterrcnn}, whose result is shown in Tab.~\ref{Tab:assinger} d. 
It shows that the performance will drop by a large margin compared with Tab.~\ref{Tab:assinger} c., which is because our proposed denoising process requires deleting over-generated bounding boxes in 3D. 
The max-IoU based assignment treats all qualified proposals as positive. Therefore the predicted score of max-IoU based models cannot be used as removing unnecessary boxes.

\noindent \textbf{Study of Different Intra-Proposal Attention in MonoXiver Structure:} In the main paper, we first fuse the projection-point appearance encoding $\mathbf{f}^{pt}_{9\times N\times d}$ and the geometric encoding $\mathbf{f}^{geo}_{g\times N\times d}$, and then we decode the RoI appearance encoding $\mathbf{f}^{roi}_{1\times N\times d}$ with the 9-token geometry-aware projection-point encoding in another cross-attention module. For convenience, we denote the first fusion stage as the encoder stage and the second stage as the decoder stage. We report more study results on fusion structures by changing query inputs at the encoder and decoder stages. The result is shown in Tab.~\ref{Tab:fuse}. The result shows that using the appearance feature as decoder query input achieves better performance for easy and moderate 
instances. The reason might be that using queries as input will keep more RoI information in the cross-attention calculation process.

\begin{table} [h!]
\begin{center}

\resizebox{0.8\linewidth}{!}{
    \begin{tabular}{c|c|c|c}
    \toprule
    & Appearance & Geometry & Easy/Mod./Hard \\
    \midrule
    
    a. & Decoder Q & Encoder K,V & \textbf{30.48}/\textbf{22.40}/{19.13} \\
    
    b. & Decoder Q & Encoder Q & 30.00/22.07/18.78 \\
    
    c. & Encoder Q & Decoder Q & 29.33/21.87/19.02 \\
    
    d. & Encoder K,V & Decoder Q & 29.42/22.03/\textbf{19.18} \\

    \bottomrule
    \end{tabular}
}
\end{center}
\vspace{-3mm}
\caption{Ablation studies on different MonoXiver structures. The setting a. is used in our main experiments in the main paper.} \label{Tab:fuse}
\vspace{-3mm}
\end{table}

\subsection{Detection Performance on Other Categories} \label{sec:other}

KITTI has limited samples of other categories (pedestrians and cyclists). Their performance is empirically unstable, which is reported in \cite{dfm, gupnet}. Therefore, in the main paper, we mainly focus on the detection performance of car category. Here, we also discuss related empirical upperbound analysis and experiment results in Tab.~\ref{Tab:upper_bound} and Tab.~\ref{Tab:ped_cyc} for reference.

\begin{table}[h]
\begin{center}
\resizebox{0.46\textwidth}{!}{
    \begin{tabular}{c|c|ccc|ccc}
    \toprule
    
    \multirow{2}{*}{Range} & \multirow{2}{*}{Stride} & \multicolumn{3}{c|}{\textit{Val, $AP_{R40}$, Ped.}} & \multicolumn{3}{c}{\textit{Val, $AP_{R40}$, Cyc.}}  \\
    
    & & Easy & Moderate & Hard & Easy & Moderate & Hard \\
    \midrule
    
    \multicolumn{2}{c|}{MonoCon \cite{monocon}} & 1.46 & 1.31 & 0.99 & 7.60 & 4.35 & 3.55\\
    
    \midrule

    $\pm$ 1.5 & 0.2 & 33.75 \textcolor{blue}{$_\uparrow$32.29} & 27.06 \textcolor{blue}{$_\uparrow$25.75} & 23.09 \textcolor{blue}{$_\uparrow$22.10} & 39.09 \textcolor{blue}{$_\uparrow$31.49} & 21.70 \textcolor{blue}{$_\uparrow$17.35} & 20.20 \textcolor{blue}{$_\uparrow$16.65} \\

    $\pm$ 1.5 & 0.3 & 21.61 \textcolor{blue}{$_\uparrow$20.15} & 17.59 \textcolor{blue}{$_\uparrow$16.28} & 14.46 \textcolor{blue}{$_\uparrow$13.47} & 34.02 \textcolor{blue}{$_\uparrow$26.42} & 19.00 \textcolor{blue}{$_\uparrow$14.65} & 17.48 \textcolor{blue}{$_\uparrow$13.93} \\
    
    $\pm$ 1.5 & 0.5 & 6.56 \textcolor{blue}{$_\uparrow$5.10} & 6.07 \textcolor{blue}{$_\uparrow$4.76} &	4.75 \textcolor{blue}{$_\uparrow$3.76} & 21.31 \textcolor{blue}{$_\uparrow$13.71} & 12.12 \textcolor{blue}{$_\uparrow$7.77} & 10.92 \textcolor{blue}{$_\uparrow$7.37} \\

    $\pm$ 1.5 & 0.75 & 4.85 \textcolor{blue}{$_\uparrow$3.39}	& 3.72 \textcolor{blue}{$_\uparrow$2.41} & 3.11 \textcolor{blue}{$_\uparrow$2.12} &	12.81 \textcolor{blue}{$_\uparrow$5.21} &	6.99 \textcolor{blue}{$_\uparrow$2.64} & 6.31 \textcolor{blue}{$_\uparrow$2.76} \\
    
    \bottomrule
    \end{tabular}}
\end{center}
\vspace{-3mm}
\caption{\small The empirical upper bounds of performance in Pedestrian and Cyclist on KITTI validation set based on the bottom-up anchor proposals computed by the MonoCon [19].}\label{Tab:upper_bound_ped} \vspace{-3mm}
\end{table}

\noindent \textbf{Empirical Upperbound Analysis:} As shown in Table 1 of the main paper, the empirical performance upper bound is subject to the search range and stride. Table~\ref{Tab:upper_bound_ped} shows the empirical upper bound for Pedestrians and Cyclists on the KITTI dataset. It shows that if we use a large stride and range the same as the setting used in the car category, the improvement potential is relatively small. If we use a small stride and large range, the potential improvement can be also very high.

\noindent \textbf{Experiment Results:} We report the detection performance on the Pedestrian and Cyclist category in Tab.~\ref{Tab:ped_cyc}. It shows that the MonoXiver is able to improve the detection performance on the Pedestrian category by a large margin. It has little improvement in the Cyclist category. The possible reason might be that the Cyclist category does not have enough data for MonoXiver to learn denoising over-generated Cyclist bounding boxes.

\begin{table}[h]
\begin{center}
\resizebox{0.46\textwidth}{!}{
    \begin{tabular}{c|c|ccc|ccc}
    \toprule
    
    \multirow{2}{*}{Range} & \multirow{2}{*}{Stride} & \multicolumn{3}{c|}{\textit{Val, $AP_{R40}$, Ped.}} & \multicolumn{3}{c}{\textit{Val, $AP_{R40}$, Cyc.}}  \\
    
    & & Easy & Moderate & Hard & Easy & Moderate & Hard \\
    \midrule
    
    \multicolumn{2}{c|}{MonoCon \cite{monocon}} & 1.46 & 1.31 & 0.99 & 7.60 & 4.35 & 3.55\\
    
    \midrule
    
    $\pm$ 1.5 & 0.5 & 5.59 & 4.57 & 3.64 & 6.48 & 3.45 & 2.99 \\

    $\pm$ 1.5 & 0.75 & 7.95 & 5.49 & 4.62 & 8.04 & 4.42 & 3.91 \\
    
    $\pm$ 1.5 & 1.5 & 3.57 & 2.79 & 1.90 & 7.60 & 3.87 & 3.35 \\
    
    \bottomrule
    \end{tabular}}
\end{center}
\vspace{-3mm}
\caption{\small The detection performance on Pedestrian and Cyclist on KITTI validation set.}\label{Tab:ped_cyc} \vspace{-3mm}
\end{table}

\begin{figure}[h]
    \centering
    \includegraphics[width=0.95\linewidth]{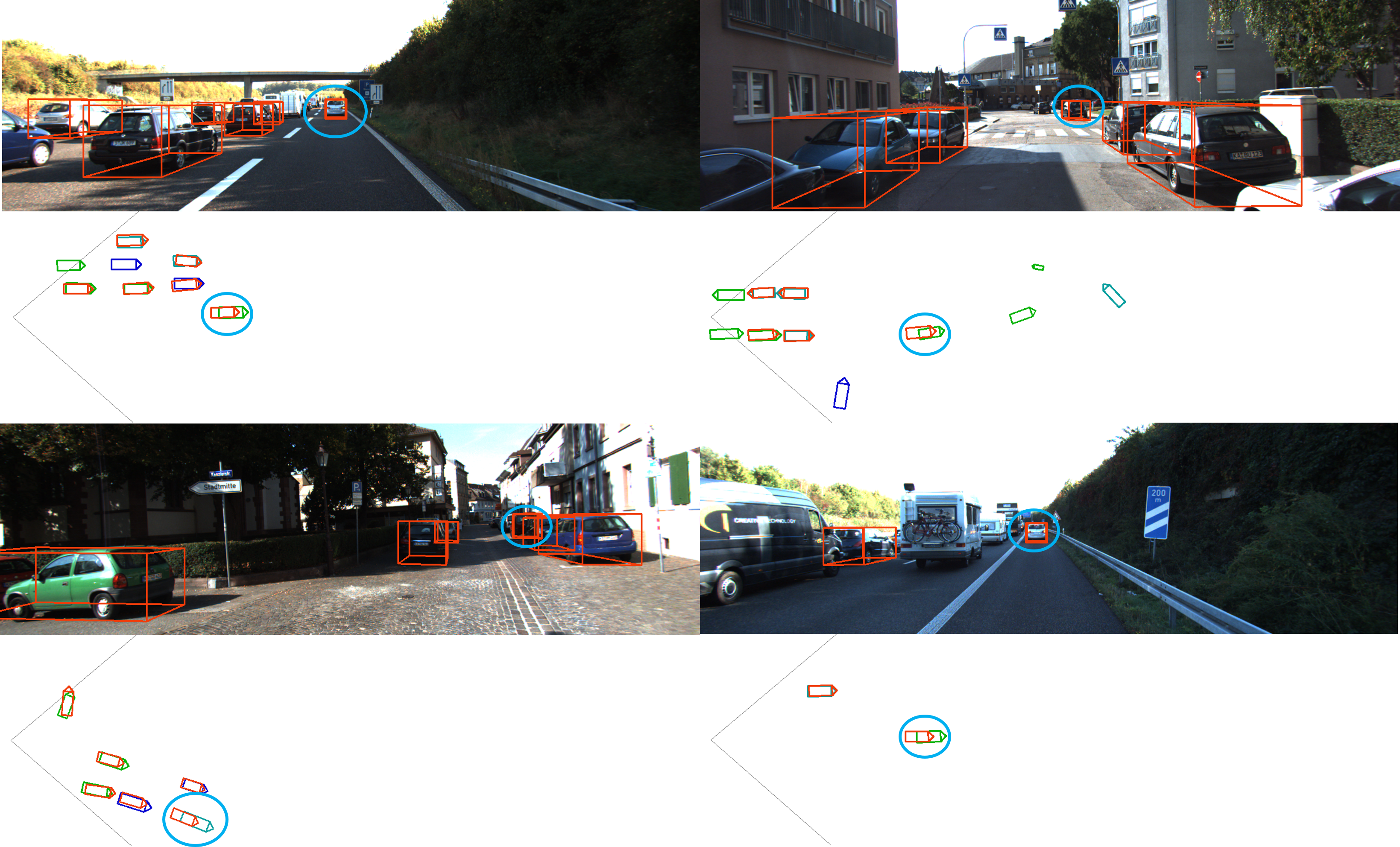}
    \caption{Qualitative results our MonoXiver with MonoCon~\cite{monocon} on KITTI \textit{validation} set \cite{mono3d}. The ground truth is shown in \textcolor{green}{green} and \textcolor{blue}{blue}. The prediction result is shown in \textcolor{red}{red}. We use top-1 prediction results for visualization.}
    \label{fig:kitti_failure}
\end{figure}

\section{Failure Case Study on KITTI} \label{sec:qual_kitti}

In Figure~\ref{fig:kitti_failure}, we present an analysis of failure cases using the baseline method MonoCon~\cite{monocon}. The results indicate that MonoXiver faces challenges in accurately classifying top-down proposals for instances that are located far away from the camera or that are directly in front of the camera. As we have discussed in the introduction of our main paper, these instances are considered to be extremely difficult negatives due to their high overlap with the ground truth, which in turn, presents an inherent challenge of depth ambiguity in monocular 3D object detection. We believe that incorporating temporal cues in our approach could be an effective solution to address this challenge, which we intend to explore in future work.

\section{Failure Case Study on Waymo} \label{sec:qual_waymo}

\begin{figure}[h]
    \centering
    \includegraphics[width=0.95\linewidth]{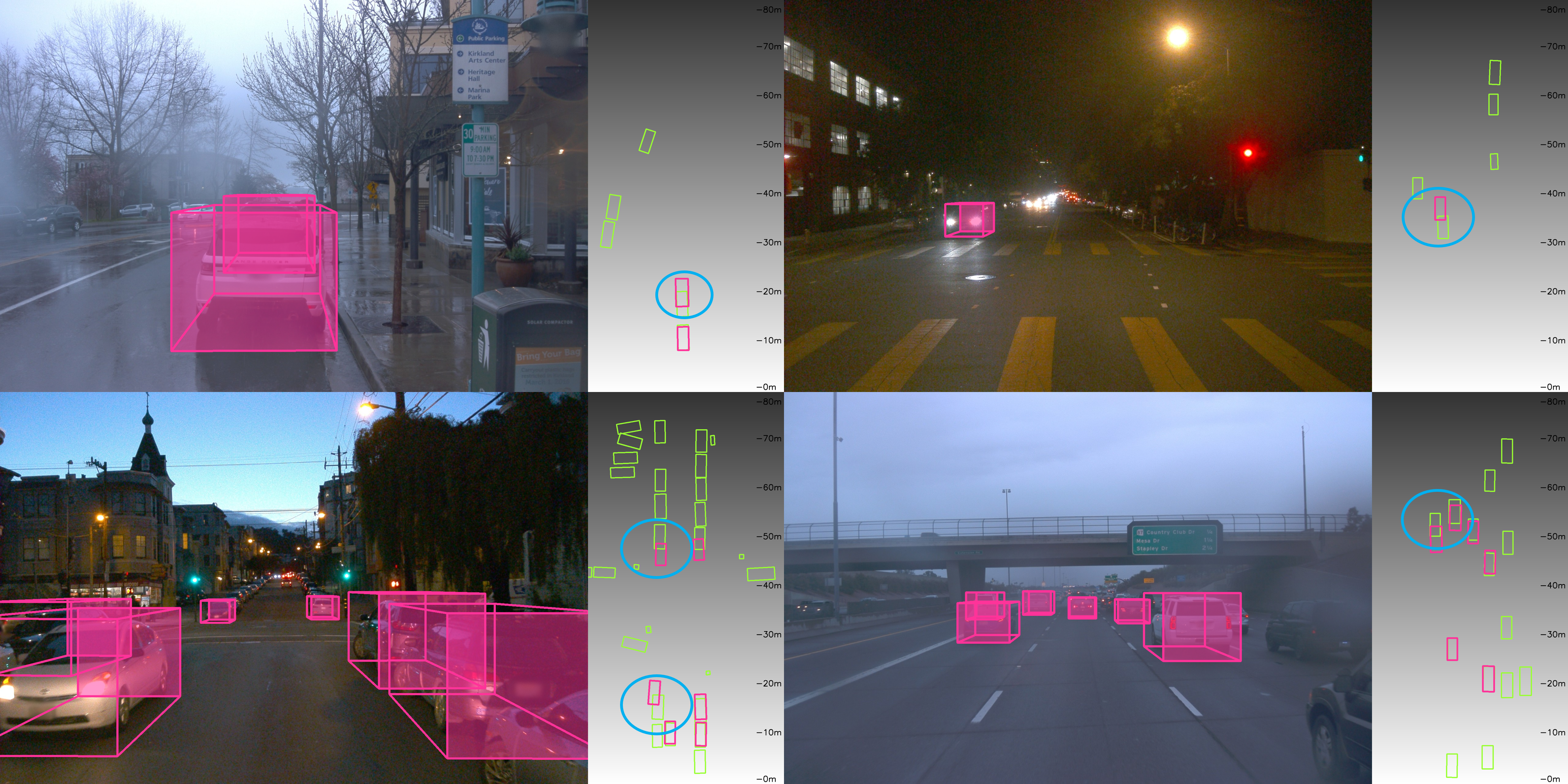}
    \caption{Qualitative results our MonoXiver with GUPNet~\cite{gupnet} on Waymo \textit{validation} set \cite{mono3d}. The ground truth is shown in \textcolor{green}{green}. The prediction result is shown in \textcolor{pink}{pink}. We use top-1 prediction results for visualization.}
    \label{fig:waymo_failure}
\end{figure}

In Figure~\ref{fig:waymo_failure}, we present an analysis of failure cases using the baseline method GUPNet~\cite{gupnet}. The results indicate that MonoXiver faces challenges in accurately classifying top-down proposals for instances that are highly occluded, truncated and that are located far from the camera. We believe that enhancing semantic cues (e.g. using spatial attention modules, larger/more powerful pretrained feature extraction backbone networks, etc.) will help resolve the occlusion and truncation issues.

\section{Detailed Network Architecture} \label{sec:network}

\noindent \textbf{Embedding MLP:} We use MLP to encode geometric features, projection point features, and RoI features. The structure is a stack of FC + LN \cite{ln} + ReLU blocks. We use one block to keep the structure simple. We use $C=256$ for embedding dimensions. \\
\noindent \textbf{Multi-head Attention layer:} We use PyTorch built-in multi-head attention for implementing intra-proposal attention and inter-proposal attention. We use 8 heads for dividing the channels. We use 2 layers of MLP (with residual connection) for projecting the attended queries. We use GELU as activate function in the MLP layer. \\
\noindent \textbf{Refinement Head:} We append two blocks of stacked MLP (FC + LN + ReLU) to the encoded queries for predicting classification scores, 3D location residuals, and 3D dimension residuals separately. We use a linear layer for prediction after the two stacked MLPs.

\end{document}